\documentclass{article}

\PassOptionsToPackage{numbers,compress}{natbib}

\usepackage[preprint]{neurips_2023}
\usepackage{caption}
\usepackage{subcaption}
\usepackage{xspace}
\usepackage{multirow}
\usepackage{graphicx}
\usepackage{lscape} 
\usepackage{rotating}
\usepackage{tabularx}
\usepackage{authblk}

\usepackage[utf8]{inputenc} 
\usepackage[T1]{fontenc}    
\usepackage{hyperref}       
\usepackage{url}            
\usepackage{booktabs}       
\usepackage{amsfonts}       
\usepackage{nicefrac}       
\usepackage{microtype}      
\usepackage{xcolor}         
\usepackage{marvosym}
\usepackage{ifsym}

\title{OpenMEDLab: An Open-source Platform for Multi-modality Foundation Models in Medicine}

\author[1]{Xiaosong Wang}
\author[1,2]{Xiaofan Zhang}
\author[3]{Guotai Wang}
\author[1]{Junjun He}
\author[4]{Zhongyu Li}
\author[5]{Wentao Zhu}
\author[7]{Yi Guo} 
\author[6]{Qi Dou}  
\author[9]{Xiaoxiao Li}
\author[1,2]{Dequan Wang}
\author[2]{Liang Hong} 
\author[8]{Qicheng Lao}
\author[10]{Tong Ruan}
\author[11]{Yukun Zhou}
\author[12]{Yixue Li}
\author[13]{Jie Zhao}
\author[1,14]{Kang Li}
\author[15]{Xin Sun}
\author[16]{Lifeng Zhu}
\author[\Letter,1]{Shaoting Zhang}
\affil[1]{Shanghai AI Laboratory, China}
\affil[2]{Shanghai Jiao Tong University, China}
\affil[3]{University of Electronic Science and Technology of China, China}
\affil[4]{Xi'an Jiaotong University, China}
\affil[5]{Zhejiang Laboratory, China}
\affil[6]{Chinese University of Hong Kong, China}
\affil[7]{Fudan University, China}
\affil[8]{Beijing University of Posts and Telecommunications, China}
\affil[9]{University of British Columbia, Canada}
\affil[10]{East China University Of Science And Technology, China}
\affil[11]{University College London, UK}
\affil[12]{Guangzhou Laboratory, China}
\affil[13]{The First Affiliated Hospital of Zhengzhou University, China}
\affil[14]{West China Hospital, Sichuan University, China}
\affil[15]{Xinhua Hospital, Shanghai Jiaotong University School of Medicine, China}
\affil[16]{Ruijin Hospital, Shanghai Jiaotong University School of Medicine, China}
\affil[\Letter]{Corresponding author(s): zhangshaoting@pjlab.org.cn}

\begin{document}

\maketitle

\begin{abstract}
  The emerging trend of advancing generalist artificial intelligence, such as GPTv4 and Gemini, has reshaped the landscape of research (academia and industry) in machine learning and many other research areas. However, domain-specific applications of such foundation models (e.g., in medicine) remain untouched or often at their very early stages. It will require an individual set of transfer learning and model adaptation techniques by further expanding and injecting these models with domain knowledge and data. The development of such technologies could be largely accelerated if the bundle of data, algorithms, and pre-trained foundation models were gathered together and open-sourced in an organized manner. In this work, we present OpenMEDLab, an open-source platform for multi-modality foundation models. It encapsulates not only solutions of pioneering attempts in prompting and fine-tuning large language and vision models for frontline clinical and bioinformatic applications but also building domain-specific foundation models with large-scale multi-modal medical data.  Importantly, it opens access to a group of pre-trained foundation models for various medical image modalities, clinical text, protein engineering, etc. Inspiring and competitive results are also demonstrated for each collected approach and model in a variety of benchmarks for downstream tasks. We welcome researchers in the field of medical artificial intelligence to continuously contribute cutting-edge methods and models to OpenMEDLab, which can be accessed via \url{https://github.com/openmedlab}.

\end{abstract}

\section{Introduction}

Recently, there has been a surge in the popularity of deep learning foundation models, particularly in the fields of natural language processing and computer vision. As a result, many milestone works have been proposed, such as Vision Transformers (ViT)~\cite{dosovitskiyimage}, Generative Pretrained Transformers (GPT)~\cite{radford2018improving}, Contrastive Language-Image Pretraining (CLIP)~\cite{radford2021learning}, and Segment Anything (SAM)~\cite{kirillov2023segment}. As the size (number of parameters) of these models grows bigger, the capacity of the models increases while the requirement of assembled data for training also inflates, following the scaling law~\cite{kaplan2020scaling}. However, for specific domains like medicine, the shortage of public availability and quality annotations has been the bottleneck for training large-scale deep learning models. Therefore, a variety of learning paradigms has been researched to overcome the roadblock besides the conventional and monotone routine of finetuning the pre-trained model (e.g., ImageNet pre-trained models~\cite{deng2009imagenet}) using domain-specific data with labels. 

One straightforward solution is to develop model adaptation techniques for medical downstream applications by leveraging the existing foundation models from other domains, especially when techniques like prompt engineering with the availability of cloud-deployed large-scale foundation models (in both highly paralleled computation and worldwide network capacity) become a reality. In this case, a foundation model, often trained with thousands of millions of multiple modalities of data, could serve as the base for building medical applications with a single or a few cases in the form of prompts. It is logically more feasible to provide such a few differentiable sample cases from a real-world scenario for the model adaptation, and it also complies with the training process of professionals in the biomedical field, e.g., medical residents and laboratory trainees.
Few-shot methods could leverage more on the distinctive representation produced by the foundation models, which has succeeded in considerable language modeling~\cite{brown2020language} and vision~\cite{dhillon2019baseline,tian2020rethinking} tasks. It also fits perfectly for the long-tailed scenario when only a few rare disease cases are available for the training.

\begin{figure}[t]
	\centering
             \includegraphics[width=\linewidth]{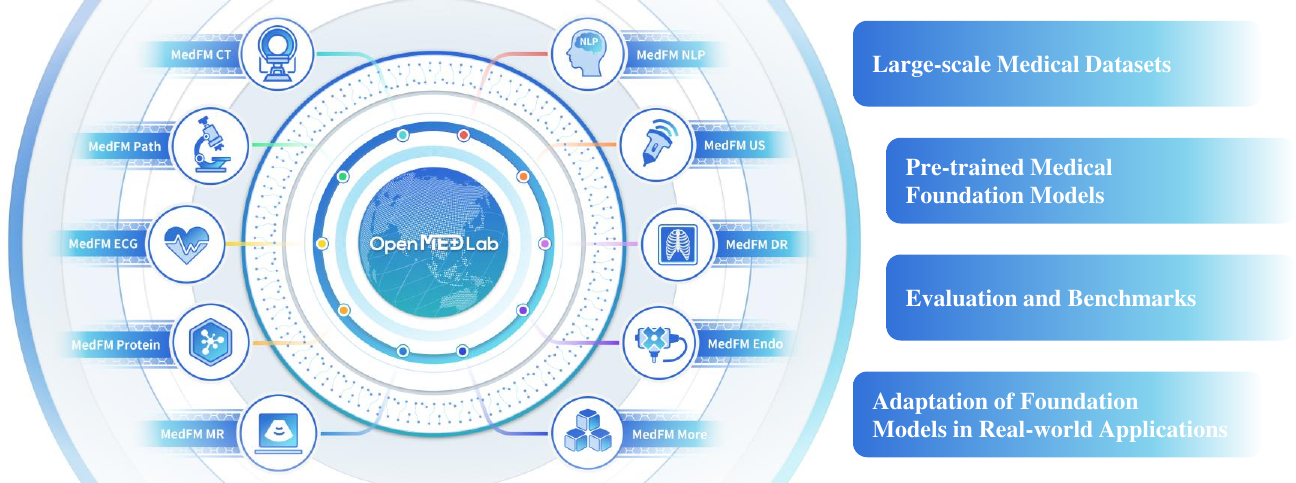}
	\caption{OpenMEDLab: an open-source platform for facilitating medical foundation models. }
	\label{fig:overview}
\end{figure}

Nevertheless, medical data could vary greatly in format, source, modality, and characteristics. Representations from the foundation models (trained with more general data, e.g., natural image and free-text natural language corpus) may not cover the domain features seamlessly. That's also why finetuning the foundation models (or pre-trained models in a supervised way, e.g., ImageNet pre-trained models) for each downstream application directly remains popular. Nonetheless, it usually employs data with labels and tunes the models in a supervised fashion. They aim to solve individual downstream tasks by utilizing the robust representation learning and generalization abilities of foundation models. One of the major drawbacks is that it remains a tedious and time-consuming job for medical professionals to hand-label volumetric data repeatedly, although attempts are made to utilize weak labels~\cite{lei2023one} and contextual information from the longitudinal data~\cite{wu2023pattern}. 
Therefore, adding an additional adaptation step could make more sense by adjusting the representation from generalist foundation models according to the domain knowledge~\cite{moor2023foundation}. 
Moreover, there are also pioneering works toward learning medical generalist/foundation models, e.g., BioGPT~\cite{luo2022biogpt}, MedSAM~\cite{ma2024segment,huang2024segment}, PLIP~\cite{huang2023visual}. This process shall require unlabeled data from the target domain, which are often much larger in quantity and relatively easier to obtain.  It will ideally have a repository to host all the models, data, and cutting-edge model training, adaptation, and benchmarking codes and showcase them as a better foundation for the various downstream clinical applications and research. 

\begin{figure}[t]
	\centering             \includegraphics[width=\linewidth]{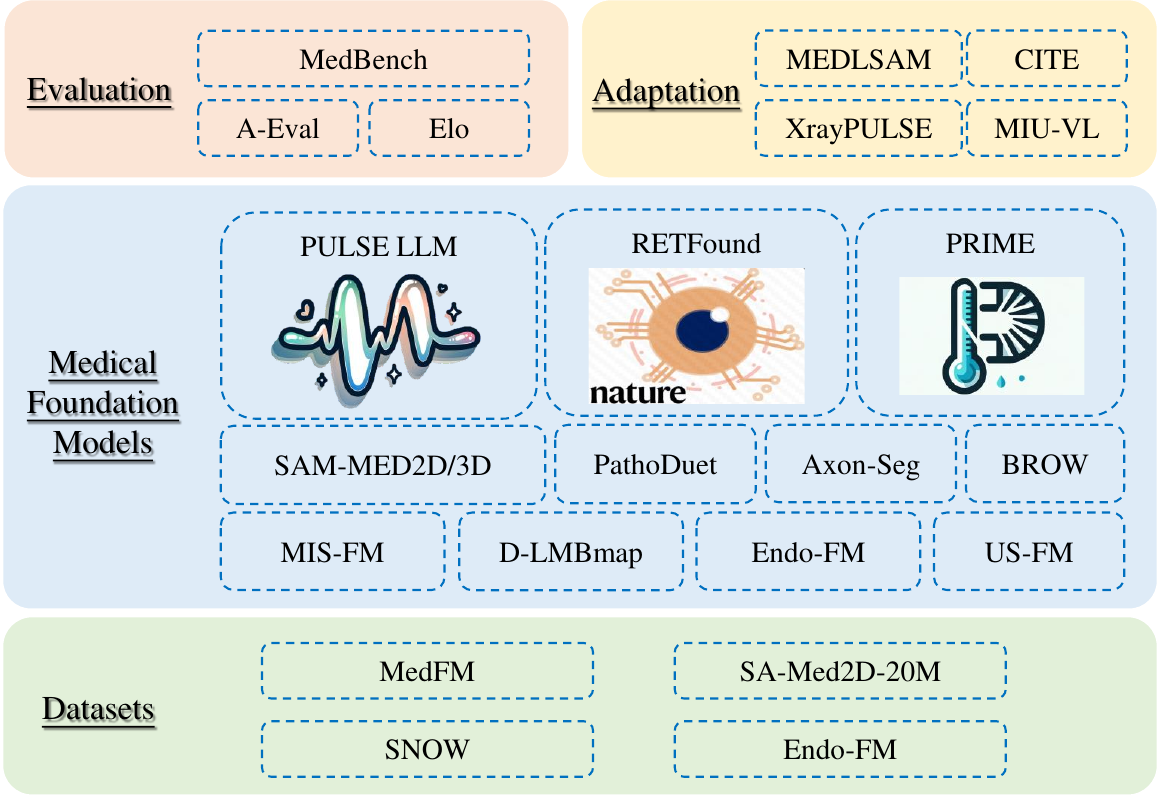}
	\caption{The overall organization of OpenMEDLab. }
	\label{fig:struct}
\end{figure}

Here, as shown in Fig.~\ref{fig:overview}, we present OpenMEDLab, which is an open-source platform to share medical foundation models in multi-modalities, e.g., medical imaging, medical NLP, bioinformatics, protein, etc. Additionally, it recruits innovative solutions in building and adapting foundation models in medicine, including the category of methodologies introduced above. Particularly, it targets promoting novel approaches to frontline clinical and research tasks in medicine, and meanwhile, it seeks solutions to achieve lower cost, higher efficiency, and better generalizability in training medical AI models. The new learning paradigm of adapting foundation models to downstream applications makes it possible to develop innovative solutions for cross-domain and cross-modality diagnostic tasks efficiently. OpenMEDLab is distinguished by several features:

\begin{itemize}
\item World's first open-source platform for medical foundation models

\item 10+ medical data modalities targeting a variety of clinical and research problems
\item Pioneering works of the new learning paradigm using foundation models, including pre-trained models, code, and data.
\item Large-scale medical datasets for pre-training and downstream applications.
\item A collaborated work from top medical institutes and facilities
\end{itemize}

We believe that the OpenMEDLab open-source platform will serve as a distinct resource and demonstration of how foundation models can be constructed and adapted for medical image analysis purposes. This will largely motivate researchers with substantial inspiration to the transition toward the novel learning paradigm of foundation models and prompt learning from the conventional transfer learning techniques. In the following sections, we will introduce the foundation models and techniques included in OpenMEDLab, i.e., prompting foundation models for medical image analysis, pre-trained medical image foundation models, medical large language models, and foundation models for various downstream clinical applications.  

\section {Overview of OpenMEDLab}
The OpenMEDLab platform provides the field with four main facilities and case studies, including Datasets, Medical Foundation Models, Evaluation, and Adaptation approaches, as illustrated in Fig.~\ref{fig:struct}. Each part covers one important stage of the lifespan of foundation models, from large-scale dataset composition to feed the foundation model training, to different pre-training paradigms in the model training, to the evaluation criterion in the usefulness of models, and finally to case study of the downstream application using foundation models. More detailed introductions of the OpenMEDLab and each part are covered in the following sections.

\section{Pretrained Medical Foundation Models}
As listed in Table~\ref{tab:models}, the OpenMEDLab comprises and publishes a variety of medical foundation models, including large Language Models (LLMs), a spectrum of imaging models focusing on different levels, and models for protein engineering. 

\begin{table}[t]
\centering
\begin{tabular}{|m{0.6in}|m{0.7in}|m{1.8in}|m{1.1in}|m{0.6in}|}

\multicolumn{5}{c}{\textbf{Pre-trained   Medical Image Foundation Models}}                                                                                                                     \\
\hline
\begin{tabular}[c]{@{}c@{}}Model\\      Category\end{tabular}                & Modality                                                            & Paper   title                                                                                                             & Dataset   for training                                                                        & \begin{tabular}[c]{@{}c@{}}Repository\\      Name\end{tabular}                                          \\
\hline
LLM                                                                          & Text                                                                & Medical   large language model                                                                                            & 20.8B   Multilingual corpus                                                                   & PULSE                                                    \\
\hline
\begin{tabular}[c]{@{}c@{}}Vision\\      Retino\end{tabular}                 & CFP/OCT                                                             & RETFound:   a foundation model for generalizable disease detection from retinal images                                    & 1.6M   CFP and OCT images                                                                     & \begin{tabular}[c]{@{}c@{}}RETFound\\ \_MAE\end{tabular} \\
\hline
\begin{tabular}[c]{@{}c@{}}Vision\\      Endo\end{tabular}                                                                  & \begin{tabular}[c]{@{}c@{}}Endoscopy   \\      videos\end{tabular}  & Foundation   Model for Endoscopy Video Analysis                                                                           & 32K   video clips (5M frames)                                                                 & Endo-FM                                                  \\
\hline
\multirow{4}{*}{\begin{tabular}[c]{@{}c@{}}Vision\\      CT/MR\end{tabular}} & CT   volumes                                                        & STU-Net:   Scalable and Transferable Medical Image Segmentation Models Empowered by   Large-Scale Supervised Pre-training & TotalSegmentator   dataset which contains 1204 images with 104 anatomical structures          & STU-Net                                                  \\
\cline{2-5}
                                                                             & CT/MR volumes                                                       & Medical   Image Segmentation Foundation Model                                                                             & 10k   public and 98k private CT volumes                                                       & MIS-FM                                                   \\
                                                                             \cline{2-5}
                                                                             & CT/MR volumes                                                       & SAM   Med2D                                                                                                               & SA   Med2D 20M: 4.6M images with 2B masks                                                     & SAM-Med2D                                                \\
                                                                             \cline{2-5}
                                                                             & CT/MR volumes                                                       & SAM-Med3D                                                                                                                 & SA   Med2D 20M: 4.6M images with 2B masks                                                     & SAM-Med3D                                                \\
                                                                             \hline
\multirow{2}{*}{\begin{tabular}[c]{@{}c@{}}Vision \\ Pathology\end{tabular}} & \begin{tabular}[c]{@{}c@{}}Pathology   \\      Images\end{tabular}  & PathoDuet:   Foundation Model for Pathological Slide Analysis of H\&E and IHC Stains                                      & 13M   patches from 11K TCGA WSIs                                                              & PathoDuet                                                \\
\cline{2-5}
                                                                             & Pathology WSI                                                       & BROWN:   Better Features for Whole Slide Image Based on Self-distillation                                                 & 10k+ WSIs without labels                                                                      & WSI\_FM                                                  \\
                                                                             \hline
\multirow{2}{*}{\begin{tabular}[c]{@{}c@{}}Vision\\      US\end{tabular}}    & Ultrasound                                                          & Deblurring   Masked Autoencoder is Better Recipe for Ultrasound Image Recognition                                         & Private   thyroid ultrasound images from 10,675 to 280,000.                                   & D-MIM                                                    \\
\cline{2-5}
                                                                             & Ultrasound                                                          & USFM:   A Universal Ultrasound Foundation Model Generalized to Tasks and Organs   towards Label Efficient Image Analysis  & Over   two million US images                                                                  & USFM                                                     \\
                                                                             \hline

\multirow{2}{*}{\begin{tabular}[c]{@{}c@{}}Vision\\      LSFM\end{tabular}}  & \begin{tabular}[c]{@{}c@{}} light-sheet \\ fluorescence \\ microscopy \end{tabular} & D-LMBmap: a fully automated deep-learning pipeline for whole-brain profiling of neural circuitry     & 100 cubes of axons and artifacts individually & Axon-Seg                                                   \\
                                                                            \hline
Protein                                                                      & \begin{tabular}[c]{@{}c@{}}Protein   \\      Sequences\end{tabular} & TemPL:   A Novel Deep Learning Model for Zero-Shot Prediction of Protein Stability and Activity                         & 96   million sequence host bacterial strain optimal growth temperatures (OGTs) and $\Delta$Tm data & PRIME                                                   \\ 
\hline
\end{tabular}
\\
\caption{Summarization of all models open-sourced in OpenMEDLab.}
\label{tab:models}
\end{table}

\subsection{Medical Large Language Models}
\begin{figure}
\centering
\includegraphics[width=0.9\textwidth]{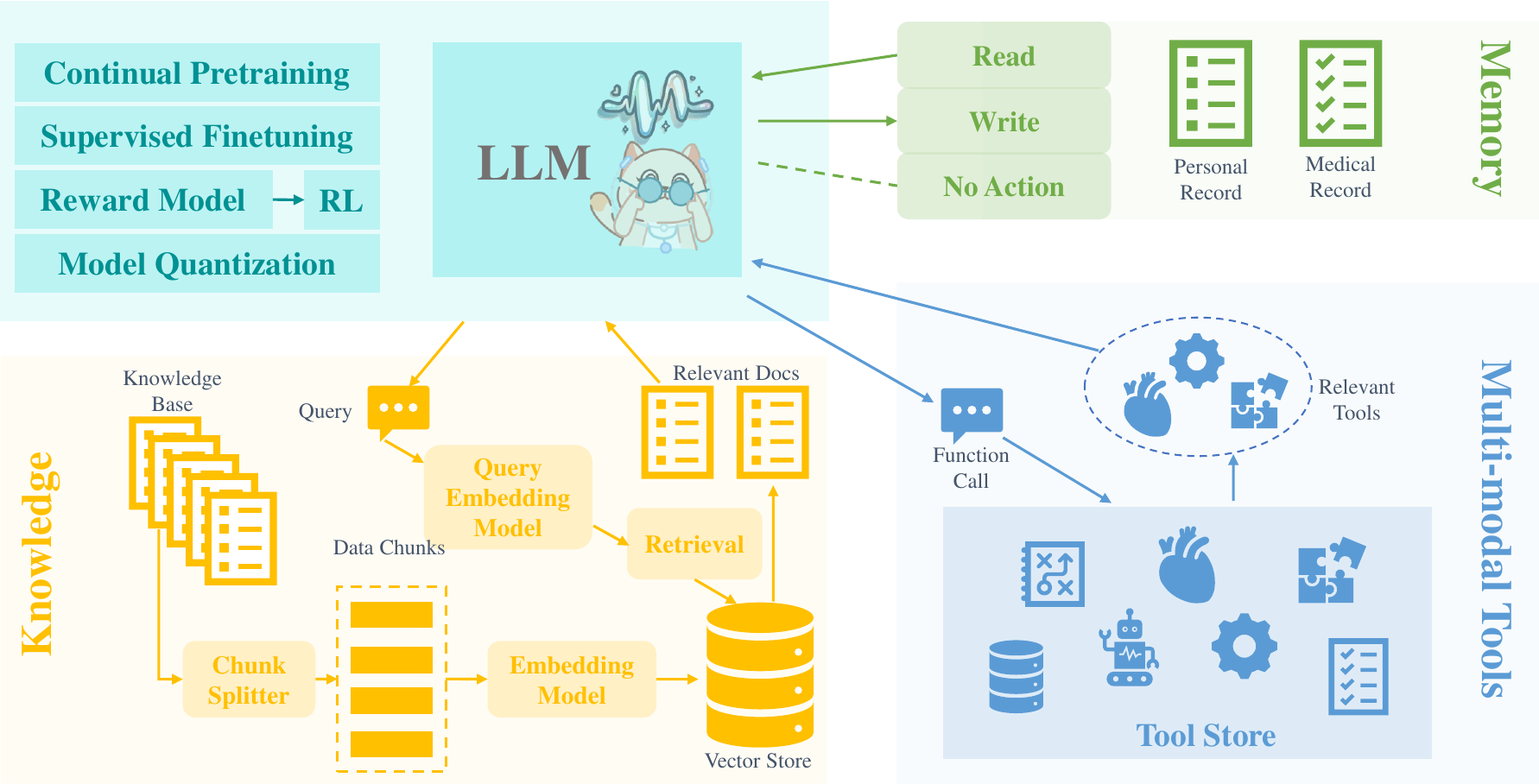}
\caption{The overview of the PULSE medical LLM.}
\label{fig:MLLM}
\end{figure}

The large medical language model, PULSE, has been released in the OpenMEDLab. Fig.~\ref{fig:MLLM} presents the overall architecture of the proposed framework. The datasets are collected from textbooks, guidelines, EHR, web Q\&A data, public instruction tuning dataset, multi-round dialog, tool-using data, etc., and further processed for continual pretraining and supervised fine-tuning. 
It also designed an online labeling tool for reinforcement learning data collection and had qualified experts from various hospital departments score and rank the responses generated by the model. 
Moreover, PULSE is trained to leverage external knowledge, long-term memory, and models from other modalities for complex downstream applications.
More details of the Pulse LLM are discussed below.

\textbf{Continual Pretraining and Supervised Finetuning.}
From the pre-trained base model to the chat model in the medical domain, we have compiled a corpus specifically tailored to this field, comprising a total of 20 billion tokens (12B for pretraining and 8B for supervised finetuning).
During the training, FlashAttention-2\cite{dao2023flashattention} and NTK-Aware Scaled RoPE\footnote{\url{https://www.reddit.com/r/LocalLLaMA/comments/14lz7j5/ntkaware_scaled_rope_allows_llama_models_to_have/}} are employed for longer context lengths.

\textbf{RLHF.}
In the next stage, we collect medical experts' feedback (ranking multiple LLM-generated responses) on the tasks in 5 areas, including medicine, operation, lab tests, radiology, nursing, and 500 diseases.
We train the reward model with self-correction techniques and implement the PPO method for RL.

\textbf{Quantization.}
In the context of expanding LLMs, considerations such as hardware costs, inference speed, and processing time in the prefill stage are becoming increasingly critical. Therefore, we employ Low-bit weight quantization methods \cite{frantar2022gptq,lin2023awq} for saving memory as well as speeding up the model in the inference stage.


\textbf{External Knowledge.}
LLMs sometimes struggle with considering issues of hallucination (prone to generating factually incorrect statements) and temporal misalignment(unable to capture the changing world), especially for knowledge-intensive tasks. Incorporating external knowledge by retrieval augmentation, i.e., Retrieval Augmented Generation(RAG), summarizing the retrieved documents, and generating the response to the question is an effective way of solving the problem.

\textbf{Long-term Memory.}
Personal information and medical records of patients could go back decades and be stored in many databases. We integrate a series of techniques like RAG, text2sql, and constrained generation with LLM to read and write the records of patients for more accurate and personalized responses. 

\textbf{Multi-modal Tools.}
After years of development and practice of AI in the medical field, there are a large number of specialized AI models, such as data quality control, vessel segmentation, lesion detection, critical judgment, etc. 
Having LLM invoke its powerful understanding, reasoning, and zero-shot capabilities to coordinate various specialized models, complex diagnostic and treatment planning problems across multiple departments and modalities now become possible.

\subsection{Pre-trained Medical Image Foundation Models}

\begin{figure}[t]
	\centering
             \includegraphics[width=\linewidth]{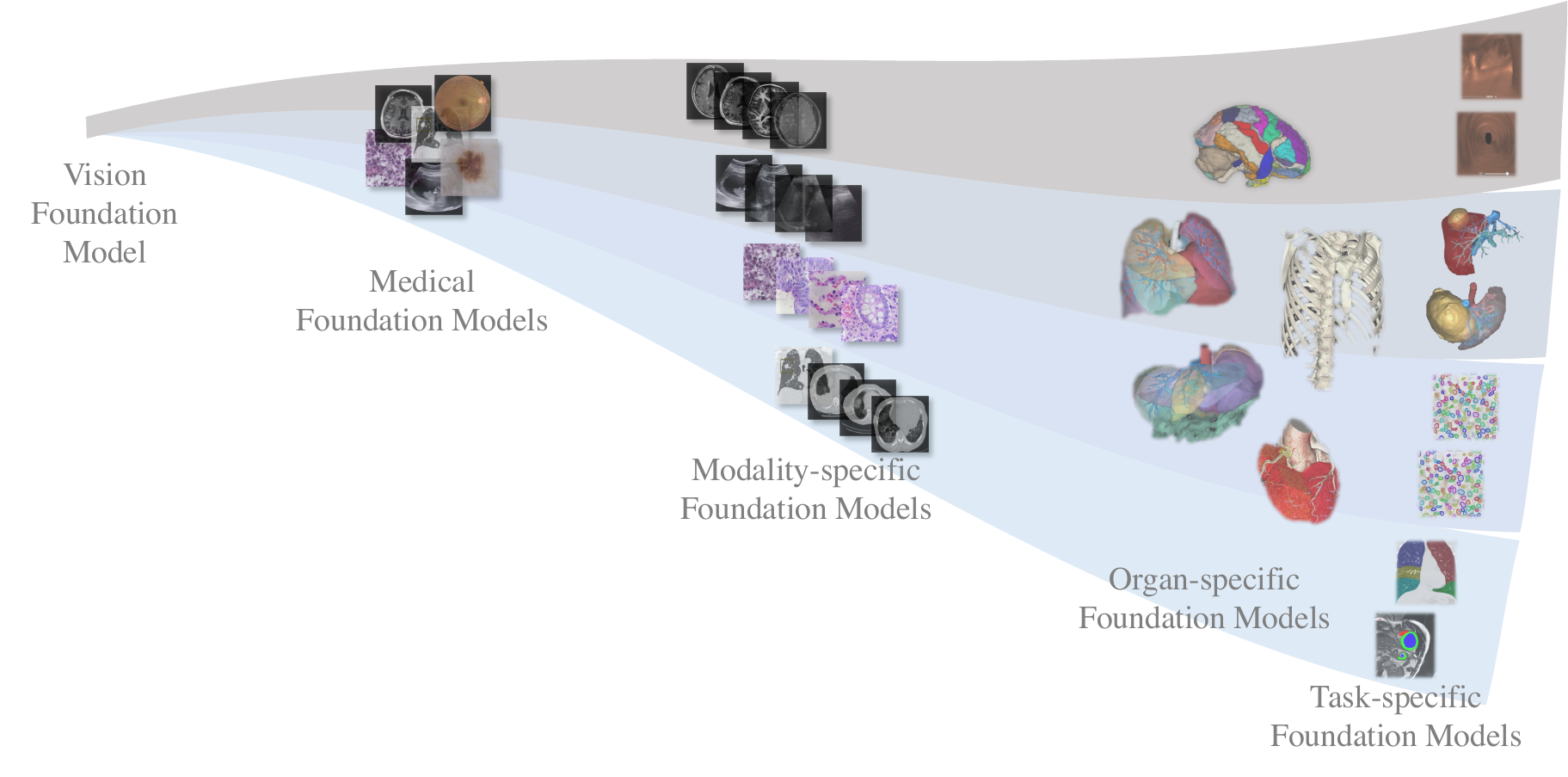}
	\caption{Different downstream applications may require foundation models with various scopes, like a spectrum of models~\cite{zhang2023challenges}. }
	\label{fig:spectrum}
\end{figure}

In this section, we introduce a bundle of medical foundation models and their applications for various imaging modalities, body parts, and more specific ones with the combination of both. In the field of medical image analysis, task-specific models are still the main approaches, especially for real-world applications such as computer-aided disease diagnosis. Developing medical foundation models presents a significant challenge due to the diverse imaging modalities used in medical imaging, as accented in~\cite{zhang2023challenges}. They could differ greatly from natural images and are based on a range of physics-based properties and energy sources, e.g., using light, electrons, lasers, X-rays, ultrasound, nuclear physics, and magnetic resonance. The produced images span multiple scales, ranging from molecules and cells to organ systems and the full body. Therefore, developing a unified multi-scale foundation model trained from a combination of these multi-modality images may be infeasible. OpenMEDLab presents a variety of foundation models and their uses in medical image analysis, ranging from modality-specific models to organ and task-specific models, as shown in Fig.~\ref{fig:spectrum}.

\textbf{RETFound\_MAE}~\cite{zhou2023foundation} Medical artificial intelligence (AI) offers great potential for recognizing signs of health conditions in retinal images and expediting the diagnosis of eye diseases and systemic disorders. However, the development of AI models requires substantial annotation, and models are usually task-specific with limited generalizability to different clinical applications. Here, we present RETFound, a foundation model for retinal images that learns generalizable representations from unlabelled retinal images and provides a basis for label-efficient model adaptation in several applications. Specifically, RETFound is trained on 1.6 million unlabelled retinal images by means of self-supervised learning and then adapted to disease detection tasks with explicit labels. We show that adapted RETFound consistently outperforms several comparison models in the diagnosis and prognosis of sight-threatening eye diseases, as well as incident prediction of complex systemic disorders such as heart failure and myocardial infarction with fewer labeled data. RETFound provides a generalizable solution to improve model performance and alleviate the annotation workload of experts to enable broad clinical AI applications from retinal imaging.

\textbf{Endo-FM}~\cite{wang2023foundation} Recent foundation models have exhibited remarkable success in various downstream tasks, such as disease diagnosis and report generation. However, a foundation model for endoscopic videos is lacking. In this work, we propose Endo-FM, a foundation model specifically designed for endoscopic video analysis. First, we build a video transformer as Endo-FM, which captures both local and global long-range dependencies across spatial and temporal dimensions. Second, we pre-train our Endo-FM using global and local views to be robust to spatial-temporal changes and discriminative across different videos. To achieve this, we construct a large-scale endoscopy video dataset by combining all publicly available datasets and a new private one. This dataset consists of over 32K video clips (5M frames), encompassing varying modalities, target organs, and disease types. Our pre-trained Endo-FM achieves promising performance on downstream tasks, surpassing state-of-the-art methods by a significant margin.

\textbf{MIS-FM}~\cite{Wang2023MisFm} Pretraining with large-scale 3D volumes has the potential for improving the segmentation performance on a target medical image dataset where the training images and annotations are limited. Due to the high cost of acquiring pixel-level segmentation annotations on the large-scale pretraining dataset, pretraining with unannotated images is highly desirable. In this work, we propose a novel self-supervised learning strategy named Volume Fusion (VF) for pretraining 3D segmentation models. It fuses several random patches from a foreground sub-volume to a background sub-volume based on a predefined set of discrete fusion coefficients and forces the model to predict the fusion coefficient of each voxel, which is formulated as a self-supervised segmentation task without manual annotations. Additionally, we propose a novel network architecture based on parallel convolution and transformer blocks that is suitable to be transferred to different downstream segmentation tasks with various scales of organs and lesions. The proposed model was pre-trained with 110k unannotated 3D CT volumes, and experiments with different downstream segmentation targets, including head, neck, thoracic, and abdominal organs, showed that our pre-trained model largely outperformed training from scratch and several state-of-the-art self-supervised training methods and segmentation models.

\textbf{STU-Net}~\cite{huang2023stu} Large-scale models pre-trained on large-scale datasets have profoundly advanced the development of deep learning. However, the state-of-the-art models for medical image segmentation are still small-scale, with their parameters only in the tens of millions. Further scaling them up to higher orders of magnitude is rarely explored. An overarching goal of exploring large-scale models is to train them on large-scale medical segmentation datasets for better transfer capacities. In this work, we design a series of Scalable and Transferable U-Net (STU-Net) models, with parameter sizes ranging from 14 million to 1.4 billion. Notably, the 1.4B STU-Net is the largest medical image segmentation model to date. Our STU-Net is based on the nnU-Net framework due to its popularity and impressive performance. We first refine the default convolutional blocks in nnU-Net to make them scalable. Then, we empirically evaluate different scaling combinations of network depth and width, discovering that it is optimal to scale model depth and width together. We train our scalable STU-Net models on a large-scale TotalSegmentator dataset and find that increasing model size brings a stronger performance gain. This observation reveals that a large model is promising in medical image segmentation. Furthermore, we evaluate the transferability of our model on 14 downstream datasets for direct inference and three datasets for further fine-tuning, covering various modalities and segmentation targets. We observe the good performance of our pre-trained model in both direct inference and fine-tuning.

\textbf{SAM-Med3D}~\cite{wang2023sammed3d} Although the Segment Anything Model (SAM) has demonstrated impressive performance in 2D natural image segmentation, its application to 3D volumetric medical images reveals significant shortcomings, namely suboptimal performance and unstable prediction, necessitating an excessive number of prompt points to attain the desired outcomes. These issues can hardly be addressed by fine-tuning SAM on medical data because the original 2D structure of SAM neglects 3D spatial information. In this paper, we introduce SAM-Med3D, the most comprehensive study to modify SAM for 3D medical images. Our approach is characterized by its comprehensiveness in two primary aspects: firstly, by comprehensively reformulating SAM to a thorough 3D architecture trained on a comprehensively processed large-scale volumetric medical dataset; and secondly, by providing a comprehensive evaluation of its performance. Specifically, we train SAM-Med3D with over 131K 3D masks and 247 categories. Our SAM-Med3D excels at capturing 3D spatial information, exhibiting competitive performance with significantly fewer prompt points than the top-performing fine-tuned SAM in the medical domain. We then evaluate its capabilities across 15 datasets and analyze it from multiple perspectives, including anatomical structures, modalities, targets, and generalization abilities. Our approach, compared with SAM, showcases pronouncedly enhanced efficiency and broad segmentation capabilities for 3D volumetric medical images.

\textbf{BROW}~\cite{wu2023brow} For pathological images, OpenMEDLab presents BROW for better features for whole slide images Based on Self-distillation. 
The model was pre-trained on a dataset containing more than 10,000 WSIs without using any labels or annotations, including about 6,000 slides from The Cancer Genome Atlas Program (TCGA), 1,000 slides from CAMELYON17, and more than 3,000 private slides. For each slide, we used CLAM to segment the tissue and exclude the blank areas, then extracted the patches within the segmented regions. It intends to produce robust and high-quality feature representations for patches and WSIs. Then, the features can be directly employed with classifiers on slide-level multi-class subtyping problems. Additionally, the trained model also performs well on patch-level classification tasks with slight fine-tuning, as demonstrated on several common downstream application datasets.

\textbf{PathoDuet}~\cite{hua2023pathoduet}Furthermore, another foundation model for histopathological image analysis is provided in OpenMEDLab, covering both H\&E and IHC stained images. The model is based on a new self-supervised learning (SSL) framework. This framework aims at exploiting characteristics of histopathological images by introducing a pretext token during the training. The pretext token is only a small piece of the images but contains special knowledge. In the task of patch positioning, the pretext token is a small patch contained in a large region. The special relation inspires us to position this patch in the region and use the features of the region to generate the patch's features from a global perspective. The patch is also sent to the encoder solely to obtain a local view feature. The two features are pulled together to strengthen the model. In the task of multi-stain reconstruction, the pretext token is a small patch cropped from an image of one stain (IHC). The main input is the image of the other stain (H\&E), which gets masked before the encoder. These two images are roughly registered, so it is possible to recover the image of the first stain, giving only a small part of itself (stain information) and a whole image of the second stain (structure information). The two parts of features are concatenated and sent into the decoder, and it finally reconstructs the image of the first stain. The model achieves outstanding performance on both linear evaluation and full finetuning of patch classification.

\textbf{D-MIM}~\cite{kang2023deblurring} Masked autoencoder (MAE) has attracted unprecedented attention and achieves remarkable performance in many vision tasks. It reconstructs random masked image patches (known as proxy tasks) during pretraining and learns meaningful semantic representations that can be transferred to downstream tasks. However, MAE has not been thoroughly explored in ultrasound imaging. In this work, we investigate the potential of MAE for ultrasound image recognition. Motivated by the unique property of ultrasound imaging in a high noise-to-signal ratio, we propose a novel deblurring MAE approach that incorporates deblurring into the proxy task during retraining. The addition of deblurring facilitates the pretraining to recover better the subtle details presented in the ultrasound images, thus improving the performance of the downstream classification task. Our experimental results demonstrate the effectiveness of our deblurring MAE, achieving state-of-the-art performance in ultrasound image classification. Overall, our work highlights the potential of MAE for ultrasound image recognition and presents a novel approach that incorporates deblurring to improve its effectiveness further.

\textbf{US-FM}~\cite{jiao2024usfm} Inadequate generality across different organs and tasks constrains the application of ultrasound (US) image analysis methods in smart healthcare. Building a universal US foundation model holds the potential to address these issues. Nevertheless, the development of such foundational models encounters intrinsic challenges in US analysis, i.e., insufficient databases, low quality, and ineffective features. In this paper, we present a universal US foundation model, named USFM, generalized to diverse tasks and organs towards label-efficient US image analysis. First, a large-scale Multiorgan, Multi-center, and Multi-device US database was built, comprehensively containing over two million US images. Organ-balanced sampling was employed for unbiased learning. Then, USFM is self-supervised and pre-trained on the sufficient US database. To extract the effective features from low-quality US images, we proposed a spatial-frequency dual-masked image modeling method. A productive spatial noise addition-recovery approach was designed to learn meaningful US information robustly,
while a novel frequency band-stop masking learning approach was also employed to extract complex, implicit grayscale distribution and textural variations. Extensive
experiments were conducted on the various tasks of segmentation, classification, and image enhancement from diverse organs and diseases. Comparisons with representative US image analysis models illustrate the universality and effectiveness of USFM. The label efficiency experiments suggest the USFM obtains robust
performance with only 20\% annotation, laying the groundwork for the rapid development of US models in clinical practices.

\textbf{Axon-Seg}~\cite{li2023d} Recent proliferation and integration of tissue-clearing methods and light-sheet fluorescence microscopy have created new opportunities to achieve mesoscale three-dimensional whole-brain connectivity mapping with exceptionally high throughput. With the rapid generation of large, high-quality imaging datasets, downstream analysis is becoming the major technical bottleneck for mesoscale connectomics. Current computational solutions are labor-intensive with limited applications because of the exhaustive manual annotation and heavily customized training. Meanwhile, whole-brain data analysis always requires combining multiple packages and secondary development by users. To address these challenges, we developed D-LMBmap, an end-to-end package providing an integrated workflow containing three modules based on deep-learning algorithms for whole-brain connectivity mapping: axon segmentation, brain region segmentation, and whole-brain registration. D-LMBmap does not require manual annotation for axon segmentation and achieves quantitative analysis of whole-brain projector in a single workflow with superior accuracy for multiple cell types in all of the modalities tested.

\subsection{Foundation Models for Protein Engineering}
OpenMEDLab also encapsulates the advances in the research field of protein engineering (see projects in Protein Engineering). As a pioneering work, we introduce TemPL~\cite{tan2023templ}, a novel deep-learning approach for zero-shot prediction of protein stability and activity, harnessing temperature-guided language modeling. By assembling an extensive dataset of 96 million sequence-host bacterial strain optimal growth temperatures (OGTs) and $\Delta Tm$ data for point mutations under consistent experimental conditions, we effectively compared TemPL with state-of-the-art models. Notably, TemPL demonstrated superior performance in predicting protein stability. An ablation study was conducted to elucidate the influence of OGT prediction and language modeling modules on TemPL's performance, revealing the importance of integrating both components. Consequently, TemPL offers considerable promise for protein engineering applications, facilitating the design of mutation sequences with enhanced stability and activity.

\section{Datasets and Benchmarks}
\begin{figure}[t]
    \centering             \includegraphics[width=\linewidth]{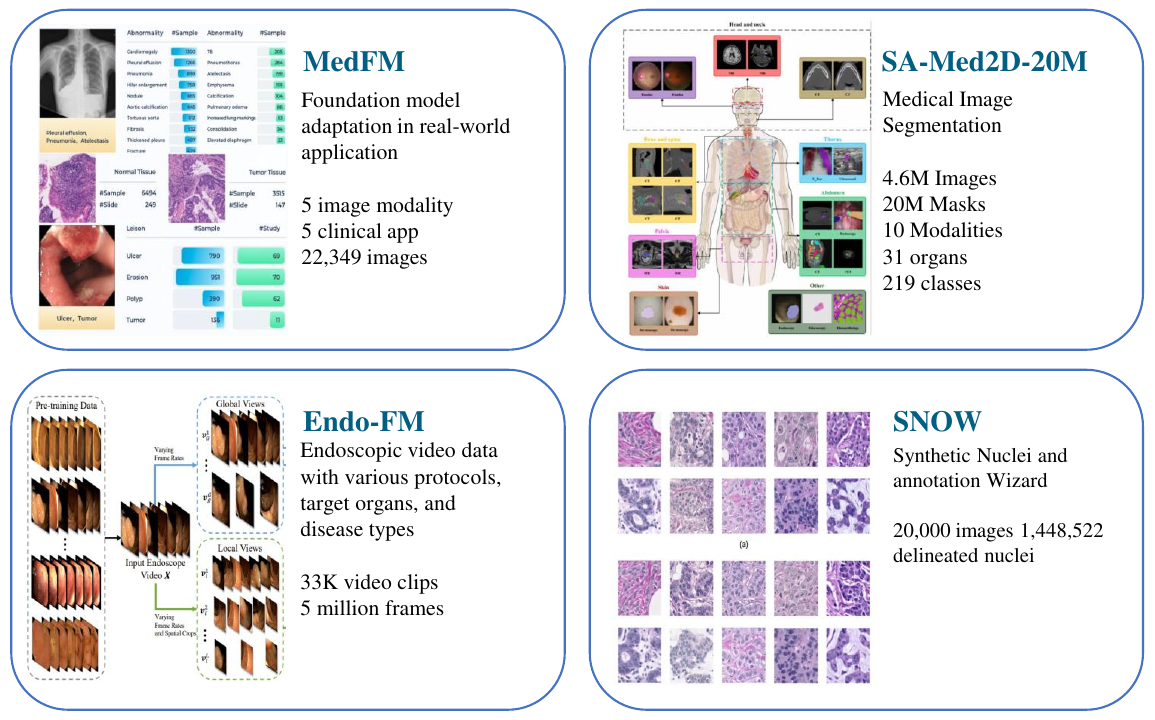}
	\caption{Four datasets in OpenMEDLab with sample images. }
	\label{fig:datasets}
\end{figure}

As an essential component in building medical foundation models, large-scale datasets are especially scarce and invaluable to make them publicly accessible. So far, OpenMEDLab has collected four large-scale datasets for the foundation model training~\cite{wang2023foundation,ye2023sa,ding2023large} and adaptation~\cite{wang2023medfmc} in the medical field. Fig.~\ref{fig:datasets} illustrates some sample images from these four datasets. In addition, OpenMEDLab continuously updates and includes some popular datasets that are previously published, e.g., WORD~\cite{luo82word} and SegRap2023~\cite{luo2023segrap2023}.

\subsection{Evaluation of Foundation Model Adaptation in Medical Image Anslysis}
Moreover, there is also an urgent demand to set up datasets and benchmarks to promote innovation in this fast-marching research field and properly evaluate the performance gain and other cost-effective aspects. There are benchmarks~\cite{sun2020fewshot,shakeri2022fhist} for the few-shot learning tasks. Nonetheless, they focus more on each individual data modality and task. Here, we will instead promote the generalizability of the few-shot learning methods, i.e., strengthening their overall performance on various data modalities and tasks. Five new sets of medical imaging data~\cite{dataset_medfmc} from multiple institutes targeting a variety of real-world clinical tasks (22,349 images in total), i.e., thoracic diseases screening in X-rays, pathological lesion tissue screening, lesion detection in endoscopy images, neonatal jaundice evaluation, and diabetic retinopathy grading, which are preprocessed using tools in~\cite{duan2020sensecare}.
The proposed dataset aims to examine the generalizability of the evaluated method. The benchmarked approach should achieve an overall superior performance on all five prediction tasks, which are largely varied in data modality and image characteristics.

\subsection{Evaluation of Medical Large Language Models}

\textbf{Elo}. In order to balance costs, we primarily utilize GPT-4 as an alternative to human evaluation like ~\cite{chiang2023vicuna,dettmers2023qlora,bai2022training}. As described in \citep{dettmers2023qlora}, the comparative randomness in model comparisons solely based on GPT-4 scores is substantial. This aligns with our observations. Therefore, we have adopted the widely used Elo Rating tournament evaluation~\citep{elo1967proposed}, a method for calculating the relative skill levels of players in zero-sum games. For cost considerations, we chose to perform 360 rounds of random evaluation on each dataset. The order in which models compete against each other in the PK (player versus player) matches was randomized to counteract any order-related bias, with a random seed set to 42. The Elo rating parameters used were K=4 and an initial rating of 1000\footnote{The implementation code for the Elo rating and other hyperparameters can be referred to Github repository: \url{https://github.com/openmedlab/PULSE-EVAL}}. 
Sample results for the popular models (e.g., Huatuo~\cite{he2019applying}, BenTsao~\cite{du2023calla}, and BianQue~\cite{chen2023bianque}) are shown in Table~\ref{tab:Elo-LLM}.









\begin{table}
	\centering
        \resizebox{\textwidth}{!}{
	\begin{tabular}{l p{1cm} p{1cm} p{1cm} p{1cm} p{1cm} p{1cm} p{1cm} p{1cm} p{1cm} }
    \toprule
Model Name   &   AVG Rank &   MedQA USMLE &   MedQA Mainland &   Prompt CBLUE &   Web Med QA &   Checkup QA &   Medicine QA &   Dialog Summ &   Med Triage (F1) \\
\midrule
GPT-4        &       1.25 &          1129 &             1117 &          1110 &       1116 &        1096 &         1098 &         1109 &             0.65 \\
PULSE-pro    &       1.75 &          1089 &             1092 &          1088 &       1119 &        1105 &         1083 &         1096 &             0.63 \\
ChatGPT      &       4.00 &          1086 &             1057 &          1064 &       1053 &        1020 &         1029 &         1080 &             0.43 \\
PULSE-20B     &       4.12 &          1042 &             1024 &          1039 &       1059 &        1049 &         1069 &         1076 &             0.40 \\
Baichuan2    &       4.50 &          1024 &             1041 &          1065 &       1044 &        1062 &         1035 &         1069 &             0.33 \\
ChatGLM3     &       5.62 &          1038 &             1062 &           997 &       1012 &        1003 &         1024 &         1021 &             0.06 \\
HuatuoGPT2   &       7.62 &           955 &              993 &           985 &        963 &         983 &         1003 &          980 &             0.01 \\
QiZhenGPT    &       8.38 &           955 &              959 &           945 &        989 &        1039 &          932 &          921 &             0.00 \\
BenTsao      &       8.75 &           961 &              921 &           936 &        910 &         927 &          986 &          920 &             0.02 \\
BianQue2     &      10.12 &           913 &              928 &           919 &        988 &         974 &          900 &          908 &             0.00 \\
MING         &      10.75 &           902 &              909 &           924 &        867 &         862 &          960 &          918 &             0.01 \\
DoctorGLM    &      11.12 &           906 &              896 &           930 &        879 &         880 &          880 &          905 &             0.00 \\
    \bottomrule
    \end{tabular}}
     \caption{Evaluation results.}
	\label{tab:Elo-LLM}
\end{table}

\begin{figure}[t]
    \centering             \includegraphics[width=\linewidth]{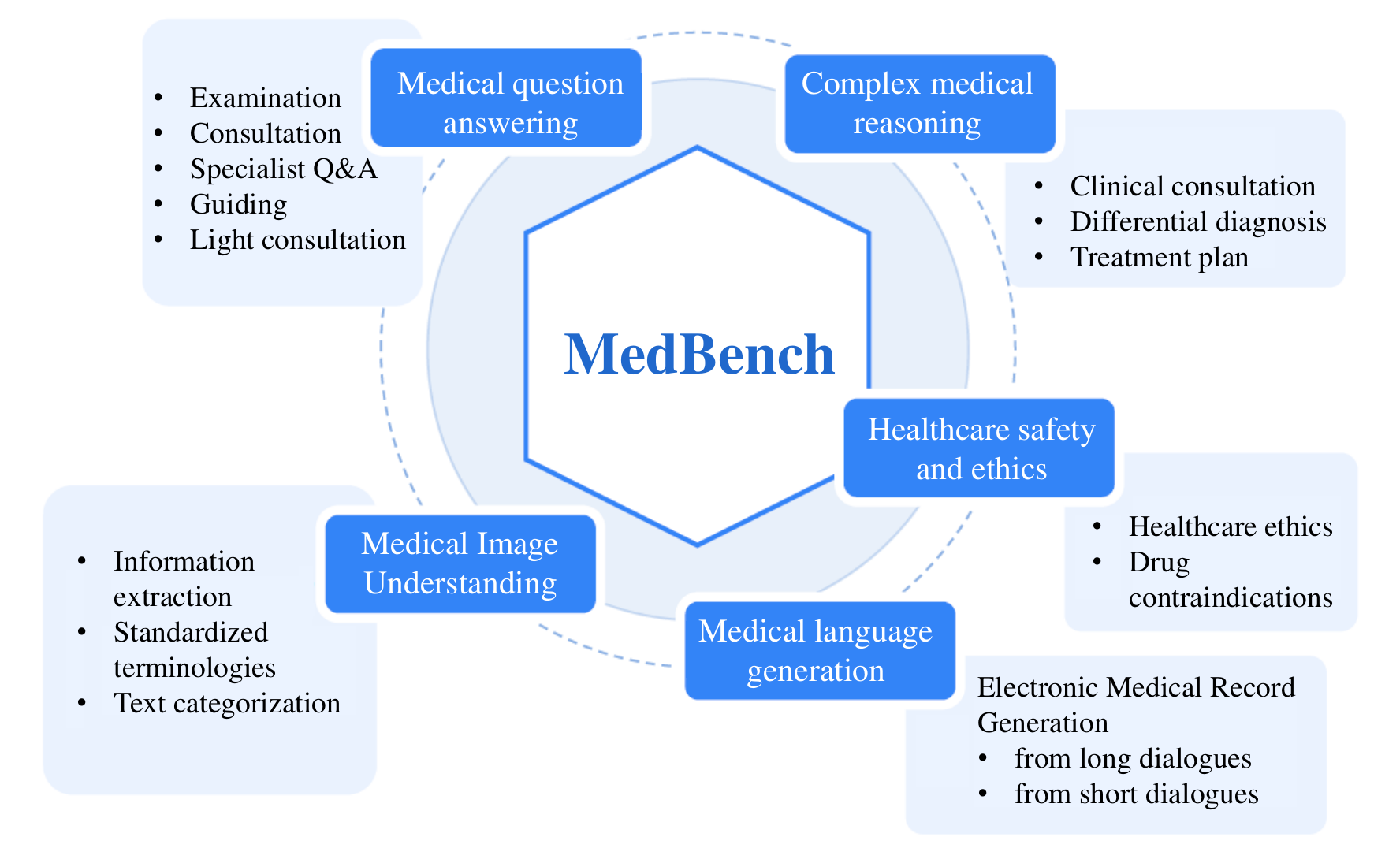}
	\caption{Overview of the MedBench evaluation metrics. }
	\label{fig:medbench}
\end{figure}

\textbf{MedBench}. 
MedBench is designed as an accessible and automatic Benchmarking System for evaluating multiple perspectives on Chinese Medical Large Language Models. Ensuring the general efficacy and goodness for human beings from medical large language models (LLM) before real-world deployment is crucial.  In MedBench, it introduces a multi-dimensional benchmarking system for Chinese medical LLM. Driven by the currently largest evaluation dataset (over 300,000 questions), MedBench assesses medical language understanding, generation, knowledge question answering, complex reasoning, and healthcare safety/ethics, as shown in Fig.~\ref{fig:medbench}. MedBench also features a fully automatic cloud-based platform with a carefully designed sampling strategy and random prompt-matching mechanism to prevent shortcut learning and answer leaking. Applying MedBench to popular general and medical LLMs demonstrates unbiased, reproducible evaluation results aligning with medical professionals' perspectives. It also establishes a significant foundation for practical applications of Chinese medical LLMs. MedBench can also be publicly accessible at https://medbench.opencompass.org.cn.

\section{Prompting Foundation Models for Medical Image Analysis}

The large-scale pre-trained vision and language models have shown remarkable domain transfer capability on natural images. However, it remains less investigated whether this capability can also apply to the medical image domain due to the unique characteristics of medical images~\cite{zhang2024data}. OpenMEDLab showcases the feasibility of transferring knowledge from pre-trained vision and language foundation models to the medical domain via well-engineered medical text prompts or building visual prompts in training. We summarise the foundation model prompting-based methods below, which tackle representative tasks in medical image analysis, i.e., image classification, disease detection, and organ/lesion segmentation in a variety of imaging modalities. 

The recent surge of foundation models in computer vision and natural language processing opens up perspectives in utilizing multi-modal clinical data to train large models with strong generalizability. Yet pathological image datasets often lack biomedical text annotation and enrichment. Guiding data-efficient image diagnosis from the use of biomedical text knowledge becomes a substantial interest. One of the representative methods is to Connect Image
and Text Embedding (CITE)~\cite{zhang2023textguided} to enhance pathological image classification. CITE injects text insights gained from language models pre-trained with a broad range of biomedical texts, leading to adapt foundation models towards pathological image understanding. Through extensive experiments on the PatchGastric stomach tumor pathological image dataset, it demonstrates that CITE achieves leading performance compared with various baselines, especially when training data is scarce. CITE offers insights into leveraging in-domain text knowledge to reinforce data-efficient pathological image classification.

As another example, MIU-VL~\cite{qin2023medical} thoroughly studies the knowledge transferability of large-scale pre-trained vision language models (VLM) to the medical domain, where it shows that well-designed medical prompts are the key to eliciting knowledge from pre-trained VLMs. By prompting with expressive attributes that are shared between domains, the VLM can carry the knowledge across domains and improve its generalization. This mechanism empowers VLMs to recognize novel objects with fewer or without image samples. Furthermore, to avoid the laborious manual designing process, MIU-VL develops three approaches for the automatic generation of medical prompts, which can inject expert-level medical knowledge and image-specific information into the prompts for fine-grained grounding. It conducts extensive experiments on thirteen different medical datasets across various modalities, showing that our well-designed prompts greatly improve the zero-shot performance compared to the default prompts, and our fine-tuned models surpass the supervised models by a significant margin.

Additionally, the Segment Anything Model (SAM) has recently emerged as a groundbreaking model in the field of image segmentation. However, both the original SAM model and its medical adaptations require slice-by-slice annotations, leading to a linear increase in annotation workload with data size. OpenMEDLab introduces MedSLAM~\cite{Lei2023medlam}, which overcomes this challenge, keeping the annotation workload constant regardless of the dataset size, significantly easing the annotation process. Firstly, we introduce a few-shot localization framework, which could localize any target anatomical part within the body. To achieve this, we develop a Localize Anything Model for 3D Medical Images (MedLAM) on a vast dataset of 14,012 CT scans with two self-supervision tasks: relative distance regression (RDR) and multi-scale similarity (MSS). Secondly, we establish a process that allows accurate segmentation by integrating MedLAM with the SAM. By annotating only six extreme points along three directions on a few templates, the proposed model can automatically identify the target anatomical area on all data slated for annotation. Subsequently, our framework generates a 2D bounding box for every slice of the image, which is then leveraged by SAM to carry out segmentation. The experiments across two 3D datasets, encompassing 38 organs, demonstrate that MedSLM matches the performance of SAM and its medical variants, all while requiring only minimal extreme point annotations for the entire dataset. Furthermore, MedLAM has the potential to be seamlessly integrated with future 3D SAM models to achieve enhanced performance.

\section{Conclusion}
In this work, the OpenMEDLab open-source platform is proposed to promote the open-source of medical foundation models and associated learning paradigms in multi-modalities, e.g., medical imaging, medical NLP, bioinformatics, protein, etc. It provides a solid foundation for this early-stage research field, as it aims at revealing the power of foundation models to ease the effort of obtaining quality annotations and target improving the classification accuracy of difficult and corner cases. It also aligns with the recent trend and success of building foundation models mentioned above for a variety of downstream applications. We believe it will be a valuable resource to open and advance the research about foundation models in medicine.


\bibliographystyle{abbrvnat}
\bibliography{ref}

\begin{thebibliography}{51}
\providecommand{\natexlab}[1]{#1}
\providecommand{\url}[1]{\texttt{#1}}
\expandafter\ifx\csname urlstyle\endcsname\relax
  \providecommand{\doi}[1]{doi: #1}\else
  \providecommand{\doi}{doi: \begingroup \urlstyle{rm}\Url}\fi

\bibitem[Bai et~al.(2022)Bai, Jones, Ndousse, Askell, Chen, DasSarma, Drain,
  Fort, Ganguli, Henighan, et~al.]{bai2022training}
Y.~Bai, A.~Jones, K.~Ndousse, A.~Askell, A.~Chen, N.~DasSarma, D.~Drain,
  S.~Fort, D.~Ganguli, T.~Henighan, et~al.
\newblock Training a helpful and harmless assistant with reinforcement learning
  from human feedback.
\newblock \emph{arXiv preprint arXiv:2204.05862}, 2022.

\bibitem[Brown et~al.(2020)Brown, Mann, Ryder, Subbiah, Kaplan, Dhariwal,
  Neelakantan, Shyam, Sastry, Askell, et~al.]{brown2020language}
T.~Brown, B.~Mann, N.~Ryder, M.~Subbiah, J.~D. Kaplan, P.~Dhariwal,
  A.~Neelakantan, P.~Shyam, G.~Sastry, A.~Askell, et~al.
\newblock Language models are few-shot learners.
\newblock \emph{Advances in neural information processing systems},
  33:\penalty0 1877--1901, 2020.

\bibitem[Chen et~al.(2023)Chen, Wang, Xing, Xu, Fang, Wang, Li, Wu, Liu, Xu,
  et~al.]{chen2023bianque}
Y.~Chen, Z.~Wang, X.~Xing, Z.~Xu, K.~Fang, J.~Wang, S.~Li, J.~Wu, Q.~Liu,
  X.~Xu, et~al.
\newblock Bianque: Balancing the questioning and suggestion ability of health
  llms with multi-turn health conversations polished by chatgpt.
\newblock \emph{arXiv preprint arXiv:2310.15896}, 2023.

\bibitem[Chiang et~al.(2023)Chiang, Li, Lin, Sheng, Wu, Zhang, Zheng, Zhuang,
  Zhuang, Gonzalez, et~al.]{chiang2023vicuna}
W.-L. Chiang, Z.~Li, Z.~Lin, Y.~Sheng, Z.~Wu, H.~Zhang, L.~Zheng, S.~Zhuang,
  Y.~Zhuang, J.~E. Gonzalez, et~al.
\newblock Vicuna: An open-source chatbot impressing gpt-4 with 90\%* chatgpt
  quality.
\newblock \emph{See https://vicuna. lmsys. org (accessed 14 April 2023)}, 2023.

\bibitem[Dao(2023)]{dao2023flashattention}
T.~Dao.
\newblock Flashattention-2: Faster attention with better parallelism and work
  partitioning.
\newblock \emph{arXiv preprint arXiv:2307.08691}, 2023.

\bibitem[Deng et~al.(2009)Deng, Dong, Socher, Li, Li, and
  Fei-Fei]{deng2009imagenet}
J.~Deng, W.~Dong, R.~Socher, L.-J. Li, K.~Li, and L.~Fei-Fei.
\newblock Imagenet: A large-scale hierarchical image database.
\newblock In \emph{Proceedings of the IEEE/CVF conference on computer vision
  and pattern recognition}, pages 248--255, 2009.

\bibitem[Dettmers et~al.(2023)Dettmers, Pagnoni, Holtzman, and
  Zettlemoyer]{dettmers2023qlora}
T.~Dettmers, A.~Pagnoni, A.~Holtzman, and L.~Zettlemoyer.
\newblock Qlora: Efficient finetuning of quantized llms.
\newblock \emph{arXiv preprint arXiv:2305.14314}, 2023.

\bibitem[Dhillon et~al.(2019)Dhillon, Chaudhari, Ravichandran, and
  Soatto]{dhillon2019baseline}
G.~S. Dhillon, P.~Chaudhari, A.~Ravichandran, and S.~Soatto.
\newblock A baseline for few-shot image classification.
\newblock \emph{arXiv preprint arXiv:1909.02729}, 2019.

\bibitem[Ding et~al.(2023)Ding, Zhou, Wang, Gevaert, Metaxas, and
  Zhang]{ding2023large}
K.~Ding, M.~Zhou, H.~Wang, O.~Gevaert, D.~Metaxas, and S.~Zhang.
\newblock A large-scale synthetic pathological dataset for deep
  learning-enabled segmentation of breast cancer.
\newblock \emph{Scientific Data}, 10\penalty0 (1):\penalty0 231, 2023.

\bibitem[Dosovitskiy et~al.(2021)Dosovitskiy, Beyer, Kolesnikov, Weissenborn,
  Zhai, Unterthiner, Dehghani, Minderer, Heigold, Gelly,
  et~al.]{dosovitskiyimage}
A.~Dosovitskiy, L.~Beyer, A.~Kolesnikov, D.~Weissenborn, X.~Zhai,
  T.~Unterthiner, M.~Dehghani, M.~Minderer, G.~Heigold, S.~Gelly, et~al.
\newblock An image is worth 16x16 words: Transformers for image recognition at
  scale.
\newblock In \emph{Proceedings of the International Conference on Learning
  Representations}, 2021.

\bibitem[Du et~al.(2023)Du, Zhao, Chen, Bai, Liu, Wu, Wang, and
  Qin]{du2023calla}
Y.~Du, S.~Zhao, Y.~Chen, R.~Bai, J.~Liu, H.~Wu, H.~Wang, and B.~Qin.
\newblock The calla dataset: Probing llms' interactive knowledge acquisition
  from chinese medical literature.
\newblock \emph{arXiv preprint arXiv:2309.04198}, 2023.

\bibitem[Duan et~al.(2020)Duan, Wang, Wang, Fu, Li, Wang, Huang, Huang, Song,
  Zhao, et~al.]{duan2020sensecare}
Q.~Duan, G.~Wang, R.~Wang, C.~Fu, X.~Li, N.~Wang, Y.~Huang, X.~Huang, T.~Song,
  L.~Zhao, et~al.
\newblock Sensecare: A research platform for medical image informatics and
  interactive 3d visualization.
\newblock \emph{arXiv preprint arXiv:2004.07031}, 2020.

\bibitem[Elo(1967)]{elo1967proposed}
A.~E. Elo.
\newblock The proposed uscf rating system, its development, theory, and
  applications.
\newblock \emph{Chess Life}, 22\penalty0 (8):\penalty0 242--247, 1967.

\bibitem[Frantar et~al.(2022)Frantar, Ashkboos, Hoefler, and
  Alistarh]{frantar2022gptq}
E.~Frantar, S.~Ashkboos, T.~Hoefler, and D.~Alistarh.
\newblock Gptq: Accurate post-training quantization for generative pre-trained
  transformers.
\newblock \emph{arXiv preprint arXiv:2210.17323}, 2022.

\bibitem[He et~al.(2019)He, Fu, and Tu]{he2019applying}
J.~He, M.~Fu, and M.~Tu.
\newblock Applying deep matching networks to chinese medical question
  answering: A study and a dataset.
\newblock \emph{BMC Medical Informatics and Decision Making}, 19\penalty0
  (2):\penalty0 52, 2019.
\newblock \doi{10.1186/s12911-019-0761-8}.

\bibitem[Hua et~al.(2023)Hua, Yan, Shen, and Zhang]{hua2023pathoduet}
S.~Hua, F.~Yan, T.~Shen, and X.~Zhang.
\newblock Pathoduet: Foundation models for pathological slide analysis of h\&e
  and ihc stains.
\newblock \emph{arXiv preprint arXiv:2312.09894}, 2023.

\bibitem[Huang et~al.(2024)Huang, Yang, Liu, Zhou, Chang, Zhou, Chen, Yu, Chen,
  Chen, et~al.]{huang2024segment}
Y.~Huang, X.~Yang, L.~Liu, H.~Zhou, A.~Chang, X.~Zhou, R.~Chen, J.~Yu, J.~Chen,
  C.~Chen, et~al.
\newblock Segment anything model for medical images?
\newblock \emph{Medical Image Analysis}, 92:\penalty0 103061, 2024.

\bibitem[Huang et~al.(2023{\natexlab{a}})Huang, Bianchi, Yuksekgonul, Montine,
  and Zou]{huang2023visual}
Z.~Huang, F.~Bianchi, M.~Yuksekgonul, T.~J. Montine, and J.~Zou.
\newblock A visual--language foundation model for pathology image analysis
  using medical twitter.
\newblock \emph{Nature medicine}, 29\penalty0 (9):\penalty0 2307--2316,
  2023{\natexlab{a}}.

\bibitem[Huang et~al.(2023{\natexlab{b}})Huang, Wang, Deng, Ye, Su, Sun, He,
  Gu, Gu, Zhang, and Qiao]{huang2023stu}
Z.~Huang, H.~Wang, Z.~Deng, J.~Ye, Y.~Su, H.~Sun, J.~He, Y.~Gu, L.~Gu,
  S.~Zhang, and Y.~Qiao.
\newblock Stu-net: Scalable and transferable medical image segmentation models
  empowered by large-scale supervised pre-training.
\newblock \emph{arXiv preprint arXiv:2304.06716}, 2023{\natexlab{b}}.

\bibitem[Jiao et~al.(2024)Jiao, Zhou, Li, Xia, Huang, Huang, Wang, Zhang, Zhou,
  Wang, and Guo]{jiao2024usfm}
J.~Jiao, J.~Zhou, X.~Li, M.~Xia, Y.~Huang, L.~Huang, N.~Wang, X.~Zhang,
  S.~Zhou, Y.~Wang, and Y.~Guo.
\newblock Usfm: A universal ultrasound foundation model generalized to tasks
  and organs towards label efficient image analysis.
\newblock \emph{arXiv preprint arXiv:2401.00153}, 2024.

\bibitem[Kang et~al.(2023)Kang, Gao, Li, and Lao]{kang2023deblurring}
Q.~Kang, J.~Gao, K.~Li, and Q.~Lao.
\newblock Deblurring masked autoencoder is better recipe for ultrasound image
  recognition.
\newblock \emph{arXiv preprint arXiv:2306.08249}, 2023.

\bibitem[Kaplan et~al.(2020)Kaplan, McCandlish, Henighan, Brown, Chess, Child,
  Gray, Radford, Wu, and Amodei]{kaplan2020scaling}
J.~Kaplan, S.~McCandlish, T.~Henighan, T.~B. Brown, B.~Chess, R.~Child,
  S.~Gray, A.~Radford, J.~Wu, and D.~Amodei.
\newblock Scaling laws for neural language models.
\newblock \emph{arXiv preprint arXiv:2001.08361}, 2020.

\bibitem[Kirillov et~al.(2023)Kirillov, Mintun, Ravi, Mao, Rolland, Gustafson,
  Xiao, Whitehead, Berg, Lo, Dollár, and Girshick]{kirillov2023segment}
A.~Kirillov, E.~Mintun, N.~Ravi, H.~Mao, C.~Rolland, L.~Gustafson, T.~Xiao,
  S.~Whitehead, A.~C. Berg, W.-Y. Lo, P.~Dollár, and R.~Girshick.
\newblock Segment anything.
\newblock \emph{arXiv preprint arXiv:2304.02643}, 2023.

\bibitem[Lei et~al.(2023{\natexlab{a}})Lei, Su, Jiang, Gu, Wang, Liu, Wang,
  Zhang, and Zhang]{lei2023one}
W.~Lei, Q.~Su, T.~Jiang, R.~Gu, N.~Wang, X.~Liu, G.~Wang, X.~Zhang, and
  S.~Zhang.
\newblock One-shot weakly-supervised segmentation in 3d medical images.
\newblock \emph{IEEE Transactions on Medical Imaging}, 2023{\natexlab{a}}.

\bibitem[Lei et~al.(2023{\natexlab{b}})Lei, Wei, Zhang, Li, and
  Zhang]{Lei2023medlam}
W.~Lei, X.~Wei, X.~Zhang, K.~Li, and S.~Zhang.
\newblock Medlsam: Localize and segment anything model for 3d medical images.
\newblock \emph{arXiv preprint arXiv:2306.14752}, 2023{\natexlab{b}}.

\bibitem[Li et~al.(2023)Li, Shang, Liu, Zhen, Zhu, Zhong, Sturgess, Zhou, Hu,
  Zhao, et~al.]{li2023d}
Z.~Li, Z.~Shang, J.~Liu, H.~Zhen, E.~Zhu, S.~Zhong, R.~N. Sturgess, Y.~Zhou,
  X.~Hu, X.~Zhao, et~al.
\newblock D-lmbmap: a fully automated deep-learning pipeline for whole-brain
  profiling of neural circuitry.
\newblock \emph{Nature Methods}, pages 1--12, 2023.

\bibitem[Lin et~al.(2023)Lin, Tang, Tang, Yang, Dang, and Han]{lin2023awq}
J.~Lin, J.~Tang, H.~Tang, S.~Yang, X.~Dang, and S.~Han.
\newblock Awq: Activation-aware weight quantization for llm compression and
  acceleration.
\newblock \emph{arXiv preprint arXiv:2306.00978}, 2023.

\bibitem[Luo et~al.(2022{\natexlab{a}})Luo, Sun, Xia, Qin, Zhang, Poon, and
  Liu]{luo2022biogpt}
R.~Luo, L.~Sun, Y.~Xia, T.~Qin, S.~Zhang, H.~Poon, and T.-Y. Liu.
\newblock Biogpt: generative pre-trained transformer for biomedical text
  generation and mining.
\newblock \emph{Briefings in Bioinformatics}, 23\penalty0 (6):\penalty0
  bbac409, 2022{\natexlab{a}}.

\bibitem[Luo et~al.(2022{\natexlab{b}})Luo, Liao, Xiao, Chen, Song, Zhang, Li,
  Metaxas, Wang, and Zhang]{luo82word}
X.~Luo, W.~Liao, J.~Xiao, J.~Chen, T.~Song, X.~Zhang, K.~Li, D.~N. Metaxas,
  G.~Wang, and S.~Zhang.
\newblock Word: A large scale dataset, benchmark and clinical applicable study
  for abdominal organ segmentation from ct image.
\newblock \emph{Medical image analysis}, 82:\penalty0 102642,
  2022{\natexlab{b}}.

\bibitem[Luo et~al.(2023)Luo, Fu, Zhong, Liu, Han, Astaraki, Bendazzoli,
  Toma-Dasu, Ye, Chen, et~al.]{luo2023segrap2023}
X.~Luo, J.~Fu, Y.~Zhong, S.~Liu, B.~Han, M.~Astaraki, S.~Bendazzoli,
  I.~Toma-Dasu, Y.~Ye, Z.~Chen, et~al.
\newblock Segrap2023: A benchmark of organs-at-risk and gross tumor volume
  segmentation for radiotherapy planning of nasopharyngeal carcinoma.
\newblock \emph{arXiv preprint arXiv:2312.09576}, 2023.

\bibitem[Ma et~al.(2024)Ma, He, Li, Han, You, and Wang]{ma2024segment}
J.~Ma, Y.~He, F.~Li, L.~Han, C.~You, and B.~Wang.
\newblock Segment anything in medical images.
\newblock \emph{Nature Communications}, 15\penalty0 (1):\penalty0 654, 2024.

\bibitem[Moor et~al.(2023)Moor, Banerjee, Abad, Krumholz, Leskovec, Topol, and
  Rajpurkar]{moor2023foundation}
M.~Moor, O.~Banerjee, Z.~S.~H. Abad, H.~M. Krumholz, J.~Leskovec, E.~J. Topol,
  and P.~Rajpurkar.
\newblock Foundation models for generalist medical artificial intelligence.
\newblock \emph{Nature}, 616\penalty0 (7956):\penalty0 259--265, 2023.

\bibitem[Qin et~al.(2022)Qin, Yi, Lao, and Li]{qin2023medical}
Z.~Qin, H.~Yi, Q.~Lao, and K.~Li.
\newblock Medical image understanding with pretrained vision language models: A
  comprehensive study.
\newblock In \emph{Proceedings of the International Conference on Learning
  Representations}, 2022.

\bibitem[Radford et~al.(2018)Radford, Narasimhan, Salimans, Sutskever,
  et~al.]{radford2018improving}
A.~Radford, K.~Narasimhan, T.~Salimans, I.~Sutskever, et~al.
\newblock Improving language understanding by generative pre-training.
\newblock \emph{OpenAI}, 2018.

\bibitem[Radford et~al.(2021)Radford, Kim, Hallacy, Ramesh, Goh, Agarwal,
  Sastry, Askell, Mishkin, Clark, et~al.]{radford2021learning}
A.~Radford, J.~W. Kim, C.~Hallacy, A.~Ramesh, G.~Goh, S.~Agarwal, G.~Sastry,
  A.~Askell, P.~Mishkin, J.~Clark, et~al.
\newblock Learning transferable visual models from natural language
  supervision.
\newblock In \emph{Proceedings of the International conference on machine
  learning}, pages 8748--8763, 2021.

\bibitem[Shakeri et~al.(2022)Shakeri, Boudiaf, Mohammadi, Sheth, Havaei, Ayed,
  and Kahou]{shakeri2022fhist}
F.~Shakeri, M.~Boudiaf, S.~Mohammadi, I.~Sheth, M.~Havaei, I.~B. Ayed, and
  S.~E. Kahou.
\newblock Fhist: A benchmark for few-shot classification of histological
  images.
\newblock \emph{arXiv preprint arXiv: 2206.00092}, 2022.

\bibitem[Sun et~al.(2020)Sun, Li, Ding, Huang, Wang, and Yu]{sun2020fewshot}
L.~Sun, C.~Li, X.~Ding, Y.~Huang, G.~Wang, and Y.~Yu.
\newblock Few-shot medical image segmentation using a global correlation
  network with discriminative embedding.
\newblock \emph{arXiv preprint arXiv:2012.05440}, 2020.

\bibitem[Tan et~al.(2023)Tan, Li, Zhang, Hu, and Hong]{tan2023templ}
P.~Tan, M.~Li, L.~Zhang, Z.~Hu, and L.~Hong.
\newblock Templ: A novel deep learning model for zero-shot prediction of
  protein stability and activity based on temperature-guided language modeling.
\newblock \emph{arXiv preprint arXiv:2304.03780}, 2023.

\bibitem[Tian et~al.(2020)Tian, Wang, Krishnan, Tenenbaum, and
  Isola]{tian2020rethinking}
Y.~Tian, Y.~Wang, D.~Krishnan, J.~B. Tenenbaum, and P.~Isola.
\newblock Rethinking few-shot image classification: a good embedding is all you
  need?
\newblock In \emph{Proceedings of the European Conference on Computer Vision},
  pages 266--282. Springer, 2020.

\bibitem[Wang et~al.(2023{\natexlab{a}})Wang, Wang, Wang, Li, Da, Liu, Gao,
  Shen, He, Shen, Duan, Zhao, Li, Qiao, and Zhang]{dataset_medfmc}
D.~Wang, X.~Wang, L.~Wang, M.~Li, Q.~Da, X.~Liu, X.~Gao, J.~Shen, J.~He,
  T.~Shen, Q.~Duan, J.~Zhao, K.~Li, Y.~Qiao, and S.~Zhang.
\newblock A real-world dataset and benchmark for foundation model adaptation in
  medical image classification.
\newblock \emph{Scientific Data}, 10, 2023{\natexlab{a}}.

\bibitem[Wang et~al.(2023{\natexlab{b}})Wang, Wang, Wang, Li, Da, Liu, Gao,
  Shen, He, Shen, et~al.]{wang2023medfmc}
D.~Wang, X.~Wang, L.~Wang, M.~Li, Q.~Da, X.~Liu, X.~Gao, J.~Shen, J.~He,
  T.~Shen, et~al.
\newblock Medfmc: A real-world dataset and benchmark for foundation model
  adaptation in medical image classification.
\newblock \emph{Scientific Data}, 2023{\natexlab{b}}.

\bibitem[Wang et~al.(2023{\natexlab{c}})Wang, Wu, Luo, Liu, Li, and
  Zhang]{Wang2023MisFm}
G.~Wang, J.~Wu, X.~Luo, X.~Liu, K.~Li, and S.~Zhang.
\newblock Mis-fm: 3d medical image segmentation using foundation models
  pretrained on a large-scale unannotated dataset.
\newblock \emph{arXiv preprint arXiv:2306.16925}, 2023{\natexlab{c}}.

\bibitem[Wang et~al.(2023{\natexlab{d}})Wang, Guo, Ye, Deng, Cheng, Li, Chen,
  Su, Huang, Shen, Fu, Zhang, He, and Qiao]{wang2023sammed3d}
H.~Wang, S.~Guo, J.~Ye, Z.~Deng, J.~Cheng, T.~Li, J.~Chen, Y.~Su, Z.~Huang,
  Y.~Shen, B.~Fu, S.~Zhang, J.~He, and Y.~Qiao.
\newblock Sam-med3d.
\newblock \emph{arXiv preprint arXiv:2310.15161}, 2023{\natexlab{d}}.

\bibitem[Wang et~al.(2023{\natexlab{e}})Wang, Liu, Zhang, and
  Dou]{wang2023foundation}
Z.~Wang, C.~Liu, S.~Zhang, and Q.~Dou.
\newblock Foundation model for endoscopy video analysis via large-scale
  self-supervised pre-train.
\newblock In \emph{Proceedings of the International Conference on Medical Image
  Computing and Computer-Assisted Intervention}, pages 101--111. Springer,
  2023{\natexlab{e}}.

\bibitem[Wu et~al.(2023{\natexlab{a}})Wu, Gao, Hu, and Zhang]{wu2023pattern}
L.~Wu, X.~Gao, Z.~Hu, and S.~Zhang.
\newblock Pattern-aware transformer: Hierarchical pattern propagation in
  sequential medical images.
\newblock \emph{IEEE Transactions on Medical Imaging}, 2023{\natexlab{a}}.

\bibitem[Wu et~al.(2023{\natexlab{b}})Wu, Li, Du, and Zhu]{wu2023brow}
Y.~Wu, S.~Li, Z.~Du, and W.~Zhu.
\newblock Brow: Better features for whole slide image based on
  self-distillation.
\newblock \emph{arXiv preprint arXiv:2309.08259}, 2023{\natexlab{b}}.

\bibitem[Ye et~al.(2023)Ye, Cheng, Chen, Deng, Li, Wang, Su, Huang, Chen,
  Jiang, et~al.]{ye2023sa}
J.~Ye, J.~Cheng, J.~Chen, Z.~Deng, T.~Li, H.~Wang, Y.~Su, Z.~Huang, J.~Chen,
  L.~Jiang, et~al.
\newblock Sa-med2d-20m dataset: Segment anything in 2d medical imaging with 20
  million masks.
\newblock \emph{arXiv preprint arXiv:2311.11969}, 2023.

\bibitem[Zhang and Metaxas(2024)]{zhang2023challenges}
S.~Zhang and D.~Metaxas.
\newblock On the challenges and perspectives of foundation models for medical
  image analysis.
\newblock \emph{Medical Image Analysis}, 91:\penalty0 102996, 2024.

\bibitem[Zhang et~al.(2023)Zhang, Gao, Zhou, Wang, Qiao, Zhang, and
  Wang]{zhang2023textguided}
Y.~Zhang, J.~Gao, M.~Zhou, X.~Wang, Y.~Qiao, S.~Zhang, and D.~Wang.
\newblock Text-guided foundation model adaptation for pathological image
  classification.
\newblock In \emph{Proceedings of the International conference on Medical Image
  Computing and Computer Assisted Intervention}, pages 272--282. Springer,
  2023.

\bibitem[Zhang et~al.(2024)Zhang, Gao, Tan, Zhou, Ding, Zhou, Zhang, and
  Wang]{zhang2024data}
Y.~Zhang, J.~Gao, Z.~Tan, L.~Zhou, K.~Ding, M.~Zhou, S.~Zhang, and D.~Wang.
\newblock Data-centric foundation models in computational healthcare: A survey.
\newblock \emph{arXiv preprint arXiv:2401.02458}, 2024.

\bibitem[Zhou et~al.(2023)Zhou, Chia, Wagner, Ayhan, Williamson, Struyven, Liu,
  Xu, Lozano, Woodward-Court, et~al.]{zhou2023foundation}
Y.~Zhou, M.~A. Chia, S.~K. Wagner, M.~S. Ayhan, D.~J. Williamson, R.~R.
  Struyven, T.~Liu, M.~Xu, M.~G. Lozano, P.~Woodward-Court, et~al.
\newblock A foundation model for generalizable disease detection from retinal
  images.
\newblock \emph{Nature}, pages 1--8, 2023.

\end{thebibliography}
\end{document}